\documentclass[10pt,twocolumn,letterpaper]{article}

\usepackage{iccv}
\usepackage{times}
\usepackage{epsfig}
\usepackage{graphicx}
\usepackage{amsmath}
\usepackage{amssymb}

\usepackage{booktabs}
\usepackage{arydshln}
\usepackage{adjustbox}
\usepackage{verbatim}
\usepackage{eso-pic}
\usepackage{multirow}
\usepackage{setspace}
\usepackage{comment}
\usepackage{siunitx}
\usepackage{caption} 
\usepackage[dvipsnames]{xcolor}
\usepackage{indentfirst}
\usepackage{xcolor}
\usepackage[T1]{fontenc}

\usepackage[pagebackref=true,breaklinks=true,letterpaper=true,colorlinks,bookmarks=false]{hyperref}

\iccvfinalcopy 

\def\iccvPaperID{2} 

\ificcvfinal\fi

\begin{document}

\title{Fully Quantized Always-on Face Detector Considering Mobile Image Sensors}

\author{Haechang Lee$^{1,5,*}$, \ Wongi Jeong$^{1,*}$, \ Dongil Ryu$^{5,*}$, \ Hyunwoo Je$^{5}$, \\ 
Albert No$^{4}$, \ Kijeong Kim$^{5,\dagger}$, \ and Se Young Chun$^{1,2,3,\dagger}$ \\     \vspace{-0.5em} \\ 
$^1$Dept. of ECE, \ $^2$INMC, $^3$IPAI, \ Seoul National University, \ Republic of Korea,\\
$^4$Dept. of EEE, \ Hongik University, \ Republic of Korea,\\
$^5$SK hynix, \ Republic of Korea\\
{\tt\small \{harrylee,wg7139,sychun\}@snu.ac.kr,} \ \ {\tt\small albertno@hongik.ac.kr,} \\
{\tt\small \{dongil.ryu,hyunwoo.je,kijeong1.kim\}@sk.com}
}

\maketitle
\ificcvfinal\thispagestyle{empty}\fi

\vspace{-1em}

\begin{abstract}
Despite significant research on lightweight deep neural networks (DNNs) designed for edge devices, the current face detectors do not fully meet the requirements for ``intelligent'' CMOS image sensors (iCISs) integrated with embedded DNNs. These sensors are essential in various practical applications, such as energy-efficient mobile phones and surveillance systems with always-on capabilities. One noteworthy limitation is the absence of suitable face detectors for the always-on scenario, a crucial aspect of image sensor-level applications. These detectors must operate directly with sensor RAW data before the image signal processor (ISP) takes over. This gap poses a significant challenge in achieving optimal performance in such scenarios. Further research and development are necessary to bridge this gap and fully leverage the potential of iCIS applications. In this study, we aim to bridge the gap by exploring extremely low-bit lightweight face detectors, focusing on the always-on face detection scenario for mobile image sensor applications. To achieve this, our proposed model utilizes sensor-aware synthetic RAW inputs, simulating always-on face detection processed ``before'' the ISP chain. Our approach employs ternary (-1, 0, 1) weights for potential implementations in image sensors, resulting in a relatively simple network architecture with shallow layers and extremely low-bitwidth. Our method demonstrates reasonable face detection performance and excellent efficiency in simulation studies, offering promising possibilities for practical always-on face detectors in real-world applications.
\end{abstract}
\let\thefootnote\relax\footnotetext{$*$ Equal contribution, $\dagger$ co-corresponding authors.}
\vspace{-1em}

\section{Introduction}
\label{sec:intro}
In the realm of deep neural network (DNN) methods, certain models are well-suited for software-level implementations on edge devices~\cite{ren2015faster,he2016deep,krizhevsky2017imagenet,dosovitskiy2020image,Liu_2021_ICCV}, but they might not fully consider the hardware-level considerations required in CMOS image sensors (CISs). CIS demands solutions that efficiently utilize their computing power to prevent wastage, especially in smartphones and surveillance systems.
For lightweight DNNs designed for image sensor hardware-level real-world implementations, achieving excellent efficiency with minimal reliance on an image signal processor (ISP) and its mobile application processor (AP) is crucial. However, existing networks may still be too heavy and complex for practical implementation.
As a result, there is a growing demand for cost-saving models that use low bit-width integer arithmetic for hardware-level deployment~\cite{liang2020toward}, which has become essential for energy-efficient and high-performance applications in CIS. This has led to active research on neural network quantization for various computer vision tasks, including image classification~\cite{yang2019quantization,Zhuang_2020_CVPR,Zhaohui_2020_NeurIPS}, object detection~\cite{li2019fully, yang2019quantization}, and super-resolution~\cite{Zhaohui_2020_NeurIPS,wang2021fully, hong2022cadyq}, aligning with this trend.
 
\begin{figure}[!t]
    \centering
    \includegraphics[width=0.85\linewidth]{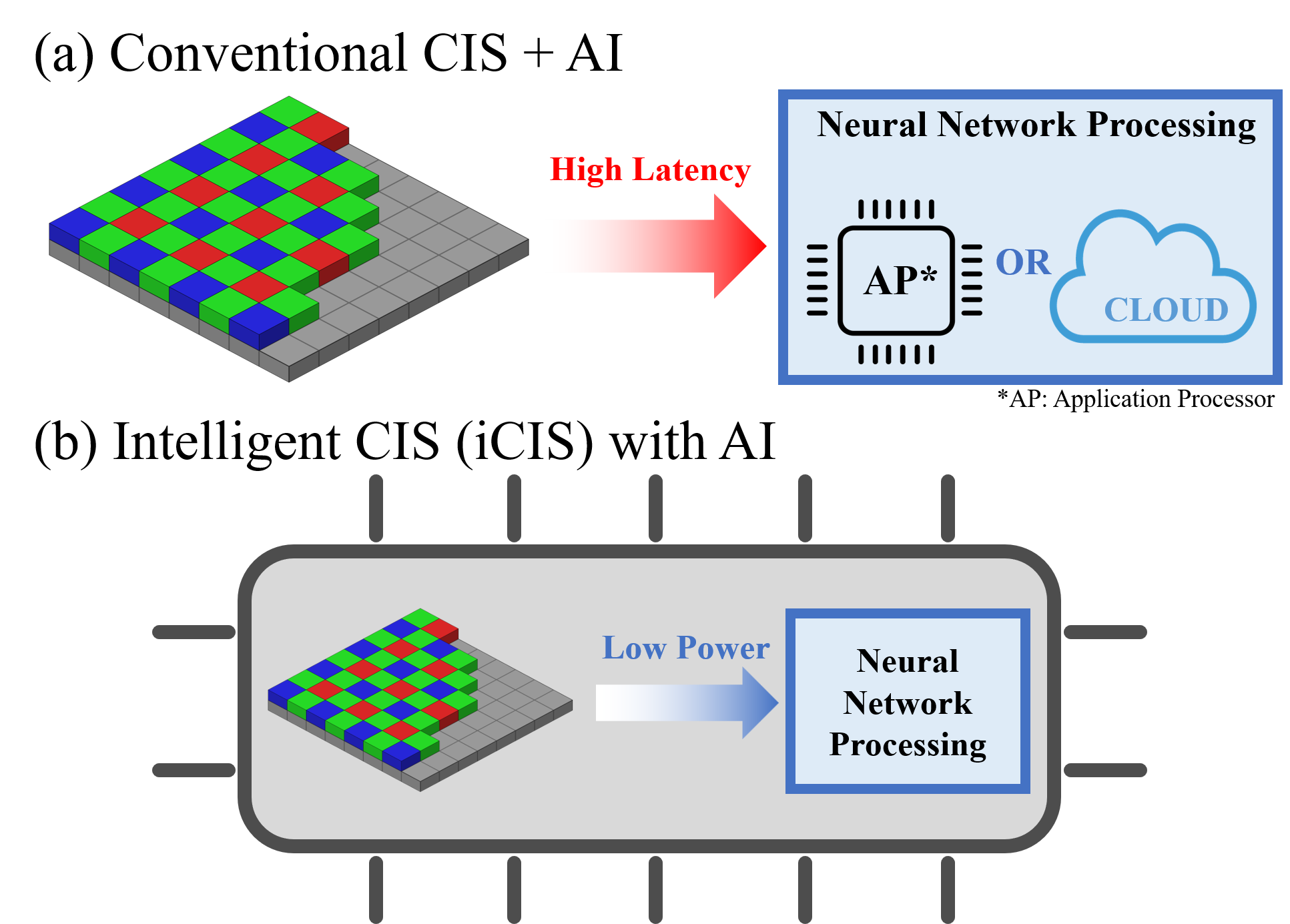}
    \vspace{-0.5em}
    \caption{The next generation of CIS; intellgent CIS (iCIS). The iCIS performs deep learning model inference directly on the sensor hardware, instead of relying on the mobile application processor (AP) or cloud server connectivity.}
    \label{fig:1_iCIS}
    \vspace{-1.5em}
\end{figure}

\noindent\textbf{Intelligent CMOS image sensor.} The CMOS image sensor (CIS), often referred to as the retina of the human eye~\cite{aw1996cmos}, is commonly used in edge devices like CCTV and smartphone cameras~\cite{el2005cmos}. The rise of intelligent CIS (iCIS) with embedded, hardware-level DNNs (as shown in Figure~\ref{fig:1_iCIS}) has gained notable attention. Utilizing accelerated MAC operations~\cite{tang2019considerations, li2020survey}, iCIS offers benefits over AP-controlled software and firmware, such as reduced power use, lower heat generation, faster inference, minimized CIS silicon area (when optimized), and less reliance on cloud-based processing~\cite{iandola2016squeezenet, alistarh2017qsgd, wang2021edge}.

\begin{figure*}[!htp]
    \centering
    \includegraphics[width=0.96\linewidth]{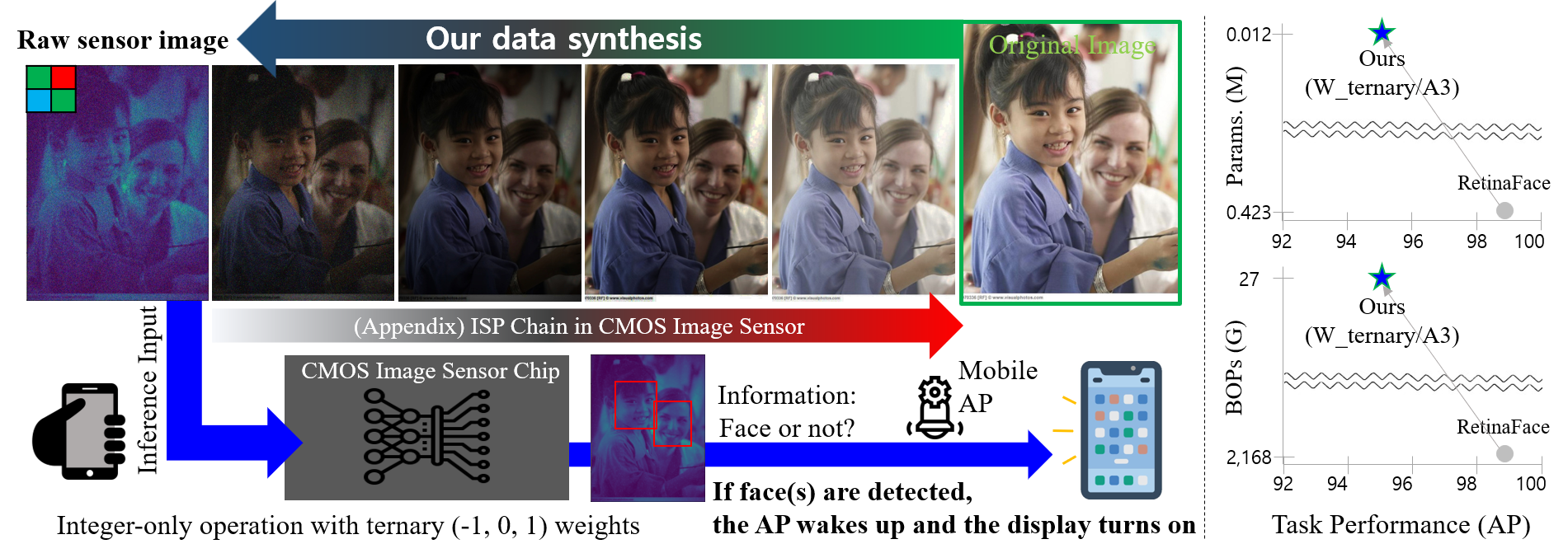}
    \caption{The upper figures illustrate our data synthesis pipeline and real AO FD application scenario. As depicted in the bottom line of the figure, the on-sensor face detector takes CIS RAW data as input. Note that the synthetic RAW inputs are actually rotated 90 degrees to the right, reflecting real CIS device characteristics, but we present the NOT rotated figures above for better visualization. Our model, TernaryFace, exhibits acceptable performance with overwhelming efficiency.}
    \vspace{-1em}
    \label{fig:2_aug_arch}
\end{figure*}

\noindent\textbf{Always-on face detection scenario.} 
Recent advances in deep learning for RAW image processing and sensor applications~\cite{ignatov2021learned, kwon2021controllable} have fueled interest in always-on (AO) face detection (FD) for mobile devices~\cite{young2019cmos, jokic2021battery, kim2019ultra, kim2020power, zhou2021heterogeneous}. The AO FD feature allows a phone's screen to activate upon facial recognition by the front camera, even if the screen is initially off, as shown in Figure~\ref{fig:2_aug_arch}. However, most existing AO FD solutions operate as software on cloud servers or firmware on ISPs, with direct on-sensor hardware implementation still under development. Performing FD at the RAW sensor level before converting to full RGB via ISP not only speeds up latency by bypassing the need for ISP or AP but also reduces memory usage by a third, as it uses 1-channel images instead of 3-channel images.

Despite the importance of face detection in practical applications, ultra-low-bit lightweight DNN-based FDs have received limited attention for edge devices, including image sensors. The strict constraints of sensor hardware for operating DNNs pose significant challenges~\cite{yu2021compute, kandaswamy2022real, iandola2016squeezenet}, rendering most FD algorithms using deep learning~\cite{earp2019face, li2019dsfd, bazarevsky2019blazeface, zhang2020refineface, chen2021yolo, facedetect-yu} unsuitable for direct execution on CIS chips.

In this paper, our focus is on developing a fully quantized ultra-low-bit face detector for on-sensor deep learning inference, specifically designed to address the limitation of previous face detectors that overlooked sensor RAW images during training and inference. Under CIS characteristics and constraints, our model outperforms previous methods in terms of efficiency under the realistic AO FD scenario. Our contributions are summarized as follows:
\begin{enumerate}
\item We present a ternary face detector tailored for real-world always-on (AO) face detection (FD) scenarios. Leveraging CIS-specialized data synthesis, a compact network architecture, and quantization techniques, our model exhibits satisfactory performance with exceptional efficiency on a RAW-like dataset, CelebCOCO, which combines CelebA~\cite{liu2015faceattributes} and MS-COCO~\cite{lin2014microsoft}.
\item We further explore ternary FD model on the WIDER FACE dataset~\cite{yang2016wider}. Our preliminary study demonstrates the initial task performance and various efficiency profiles, showcasing the foundational capabilities of our model in more challenging datasets.
\end{enumerate}

\section{Related Work}
\label{sec:relatedwork}
\subsection{Real-time Face detectors}
Viola-Jones (V-J)~\cite{viola2001rapid} is a pioneering real-time FD algorithm for edge devices. It uses Haar-like features and cascade AdaBoost algorithm, but its performance is limited~\cite{chaudhari2018face}. Most deep learning-based FD research, such as those cited~\cite{zhang2017faceboxes, chen2018mobilefacenets, martindez2019shufflefacenet, li2019airface, deng2019arcface}, focuses on efficient architectures with reasonable performance. However, these models fall short for on-sensor applications; they are trained on RGB images, are computationally heavy, and rely on floating-point arithmetic, limiting their on-sensor efficiency.

\subsection{Quantized Face Detectors}
In the case of quantized FD models, 8 bit-precision is dominant~\cite{tripathi2017lcdet, park2020real, sun2021energy, khanehgir2022light, liberatori2022yolo}. Recently, QuantFace~\cite{boutros2022quantface} introduced a 6-bit face detector. DupNet~\cite{gao2019dupnet} introduces a fully quantized face detector with totally 2-bit duplicated weights. IFQ-Net~\cite{gao2018ifq} and QMobileFaceNet~\cite{bunda2022sub} presented 2-bit FD models. LSW-Det~\cite{xu2021layer} introduced a 1-bit face detector with a combination of some real-value layers, which is not fully quantized. 

Previous studies on quantized face detection (FD) have primarily focused on simpler datasets like FDDB~\cite{fddbTech}, AgeDB~\cite{moschoglou2017agedb}, and LFW~\cite{LFWTech}. These studies provide qualitative performance metrics or detection rates, but they do not report average precision (AP), which is a more rigorous measure of performance. Furthermore, none of these studies evaluate the performance on the more challenging dataset, WIDER FACE~\cite{yang2016wider}. They also do not take into account sensor-level input data and its hardware constraints.

\section{Proposed method}
\label{sec:techniques}
In this section, we begin by introducing our proposed data synthesis techniques, which generate RAW sensor-like input images for both training and inference. These techniques aim to mimic always-on (AO) face detection (FD) in a CIS mobile application scenario. Next, we describe our model architecture, which is a significantly streamlined and modified version of RetinaFace~\cite{deng2020retinaface}. Finally, we explain how to optimally quantize our model during the training phase and outline the steps for deploying it during the inference stage.

\subsection{Data Synthesis for CIS}
\label{sec:data_synthesis}
Using 1-channel Bayer-patterned sensor RAW as the input for CIS products provides advantages, including lower hardware burden, reduced memory cost, and power savings, compared to 3-channel RGB images processed through the image signal processor (ISP).
Utilizing RAW sensor images with smaller capacity enables more efficient face detection in ultra-low-power always-on mode of cellphones without the need for the ISP to handle this task.

\subsubsection{Application-aware Data Definition}
\label{sec:data_define}
We created a synthetic dataset named CelebCOCO by combining CelebA~\cite{liu2015faceattributes} and MS-COCO~\cite{lin2014microsoft} datasets. This dataset emulates one-channel Bayer-patterned RAW sensor data, mosaiced by either the R, G, or B channels~\cite{bayer1976color}. Adding MS-COCO improves background diversity, addressing the limited variety in CelebA. This enhanced diversity is key for real-time, always-on face detection on mobile phones, as it helps reduce false positives. We used a QQVGA (160$\times$120) sensor product size, recommended for mobile CIS, for our AO FD task.
In the always-on face detection (AO FD) scenario for mobile front cameras using CIS, real faces captured are generally not too small, given the typical arm's reach of 10 to 25 inches for phone users. Therefore, during data preprocessing, we applied random cropping to the input images centered around the face, while keeping the face size ratio fixed between 20\% and 110\%. This approach aligns well with our AO FD scenario. If the cropped image exceeded the original image dimensions, we used black padding for data augmentation.
 
\subsubsection{Reversal to Raw Sensor Data}
\label{sec:reverse_raw}
Within the CIS chip, initial photon energy from objects is converted into digital RAW images, which are later processed through the ISP pipeline to obtain sRGB images. In our approach, model inference is performed on bare-sensor images before the ISP operates. Hence, we synthesize sensor-RAW-mimicking images for both the training and test sets by reversing the process, as shown in Figure~\ref{fig:2_aug_arch}.

\noindent\textbf{Sensor noise addition.} During the entire process from RAW sensor input to obtaining a high-resolution image, various types of noise can intrude the images in a sensor chip. To model this, we employ a practical mixed Poisson and Gaussian noise model~\cite{brooks2019unprocessing, luo2010research, snoeij2006cmos}:
\begin{equation}
\begin{split}
		x_n = \operatorname{Poisson}(\gamma y_n)/\gamma + \epsilon_n, \\
		\quad \epsilon \sim  \mathcal{N}(0,\sigma^2_\epsilon I),
		\quad n = 1, \ldots ,N, 
\end{split}
\end{equation}
where $y$ and $x$ are clean images and corrupted images, respectively.
The noise model consists of pixel intensity-dependent Poisson noise (caused by photon sensing) generated by $\operatorname{Poisson}(\cdot)$, where $\gamma$ is a gain parameter depending on the sensor and analog gain. Moreover , there is signal-independent Gaussian noise with standard deviation $\sigma$, and $N$ samples are considered.
We generate our train and test datasets using the parameters $\gamma = 0.01$ and $\sigma_\epsilon = 0.02$.

\noindent\textbf{Color degradation.} We adopt a color tone degradation function, inspired by~\cite{brooks2019unprocessing}, using a simple inverse smoothing curve in our synthetic process. This enables us to perform color tone degradation on open-source dataset images.

\noindent\textbf{Inverse gamma correction.} In the ISP chain, we apply gamma correction to image data to account for the non-linear perception of brightness by the human eye. For this purpose, we use a standard gamma value setting of 2.2, which is commonly used in most cameras~\cite{ershov2022ntire, peng2022bokehme}.

\noindent\textbf{Inverse lens shading correction.} Mobile camera lenses usually feature a convex center and a slight edge curvature, leading to lens shading and uneven light reception across the image. In mobile cameras with small lens sizes and high pixel counts, this effect is more pronounced, resulting in darker edges compared to the center. To address this, the ISP applies lens shading correction at the beginning of the ISP chain to ensure uniform illumination and prevent shading artifacts from affecting subsequent image processing.

\noindent\textbf{One-channel Bayer pattern.}
The ISP converts 8-bit RAW inputs to a 24-bit RGB image using interpolation (demosaicing)~\cite{zhang2019zoom}. To reverse this process, we generate a simulated Bayer~\cite{bayer1976color} color filter array (CFA) patterned sensor image, commonly found in CIS, by pixel-wise sampling from the three-channel (RGB) input. This yields a one-channel image with just one color (R, G, or B) per pixel, which serves as our input image for training, validation, and test.

\noindent\textbf{Other details.} To assess the performance of our model in real-world conditions, we utilized various data augmentation techniques. These techniques included introducing stronger Poisson noise and Salt and Pepper noise with a 50\% probability to simulate insufficient illumination and backlight environments, enhancing the model's robustness and data diversity. The data synthesis functions were applied globally, and for training, validation, and testing, we used the input images after rotating them 90 degrees to the right, accounting for the camera's installed sensor characteristic. Note that all figures and image outputs were restored by rotating them 90 degrees to the left again to ensure proper visibility throughout this paper.

\begin{figure}[!b]
    \vspace{-0.75em}
    \centering
    \includegraphics[width=0.97\linewidth]{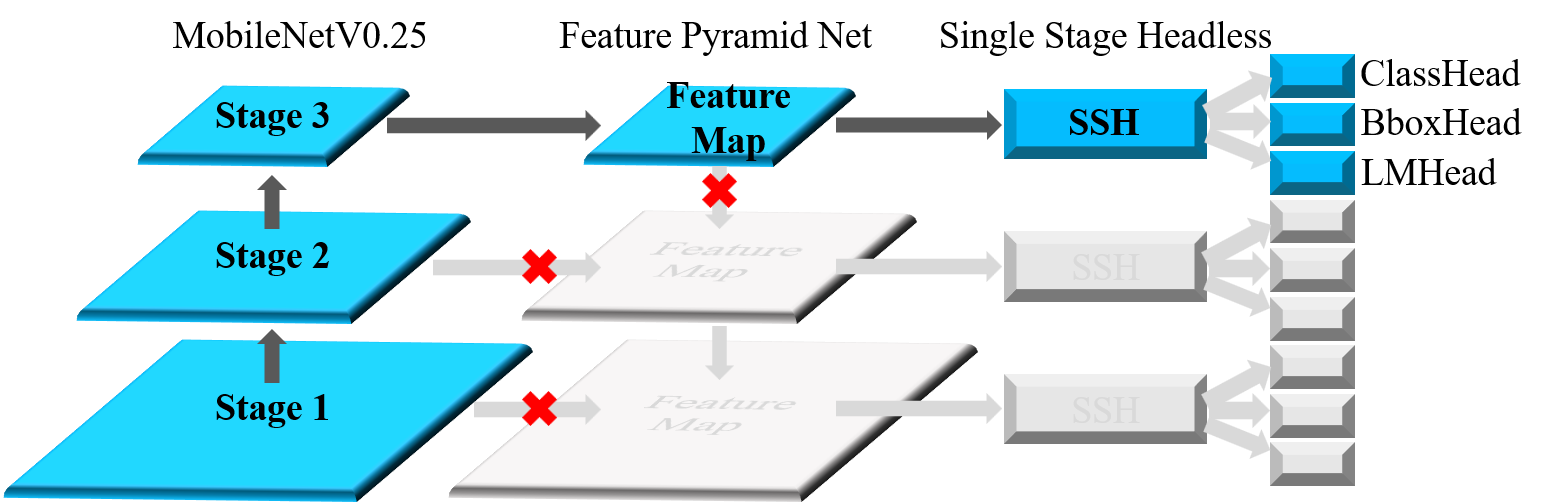}
    \vspace{-0.5em}
    \caption{Our model architecture, customized RetinaFace to make it suitable for always-on face detection considering mobile CIS applications.}
    \label{fig:3_model_arch}
\end{figure}

\subsection{Neural Network Architecture}
\label{sec:NN_arch}
We customized the RetinaFace~\cite{deng2020retinaface} model with a MobileNetV0.25 backbone, a state-of-the-art one-stage detector designed for fast inference, as shown in Figure~\ref{fig:3_model_arch}. By customizing RetinaFace, we developed a lightweight architecture specifically tailored to our CIS domain knowledge and application scenarios. The choice of MobileNet as the backbone allows us to leverage depth-wise and point-wise convolutions, minimizing CNN operations and ensuring efficiency in lightweight models. However, previous studies have shown that MobileNet models can be more sensitive to quantization~\cite{park2020profit, nagel2021white}, resulting in performance degradation compared to full-precision models like ResNet.
Considering hardware constraints, we limited the maximum input and output channels to 128 for all layers in the AI accelerator with a weight stationary structure, as exceeding this threshold can lead to latency overhead due to additional operations. To improve operational efficiency, we reduced the number of anchor boxes from over 100k to just 240 in our model.
We simplified the original RetinaFace model by replacing multiple FPN (Feature Pyramid Network) and single stage headless (SSH) modules with single feature layers and single SSH each, while preserving the capability of parallel processing at the operator level in SSH through the concatenate function. This modification was aimed at reducing the workload and computational burden while maintaining essential functionality.

\subsection{Quantization}
\label{sec:quant}

\subsubsection{Preliminary}
\label{sec:quant_pre}
Our primary focus is on extreme low-bit quantization, which is considered one of the most effective techniques for model compression.
\begin{equation}
\tilde{x} = clamp(\ \left\lfloor\frac{x}{s}\right\rceil-z\ ; q_{min}, q_{max}),
\end{equation}

\begin{equation}
q_{min}=-2^{b-1},\ q_{max}=2^{b-1}-1.
\end{equation}
\vspace{-0.9em}

\begin{figure}[!b]
    \centering
    \includegraphics[width=0.99\linewidth]{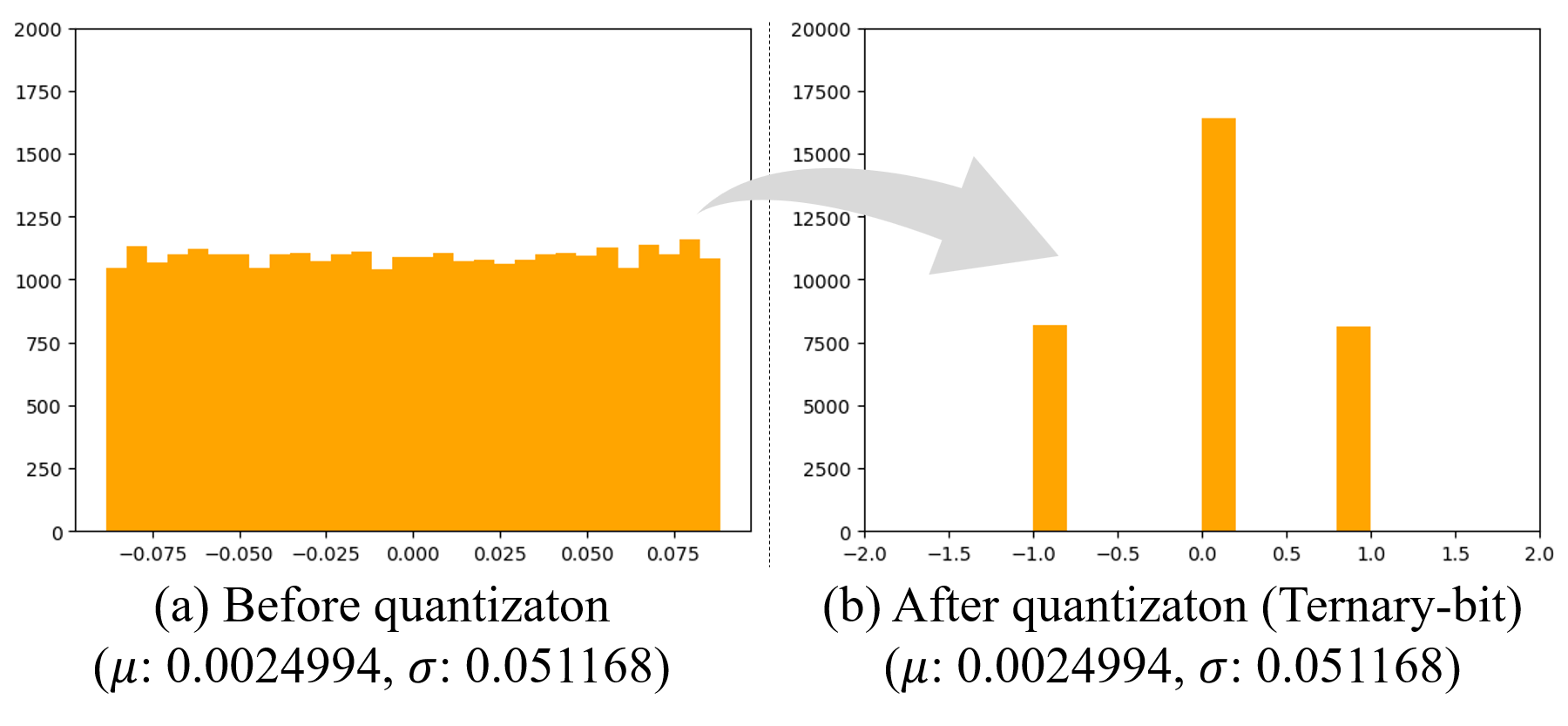}
    \vspace{-0.5em}
    \caption{The figure illustrates the changes in weights of the first convolutional layer in stage 3 of MobileNetV1 before and after quantization. Note that the weights are quantized to -1, 0, and 1 in ternary bitwidth.}
    \label{fig:4_before_after_quant}
\end{figure}

The clamp function clips the rounded-to-nearest value to be within the range of $q_{min}$ and $q_{max}$, where $z$ is a zero point, $b$ is a hyperparameter of the bit-width, and $s$ is a learnable parameter, which represents the step size. In LSQ~\cite{esser2020learned} we are implementing, the step size is considered as a learnable parameter; however, it is not commonly considered as such in general methods. Therefore, a floating-point value $x$ is quantized to integer value $\tilde{x}$ according to Eq. (2), as shown in Figure~\ref{fig:4_before_after_quant}.

\subsubsection{Quantization-aware Training}
\label{sec:qat}
Quantization-aware training (QAT) is employed to reduce quantization errors during model training, achieved by introducing fake nodes to weight/activation nodes, which induce quantization errors. This process fine-tunes the model from full precision to one that is robust to quantization errors. We then apply the target bit-depth to weights and activations after training with full precision, using straight-through estimation (STE)~\cite{bengio2013estimating} to enable gradient propagation in neural networks with threshold operations.

For our proposed model, which is a tailored version of RetinaFace, we aim to achieve strict end-to-end quantization. We apply per-tensor, uniform, and symmetric quantization as the basic policy~\cite{gholami2021survey}. Despite using a very light neural network structure with a single MobileNet~\cite{howard2017mobilenets} backbone-based Feature Layer, which is known to be sensitive to quantization, our model demonstrates acceptable performance for AO FD CIS application even when fully quantized to ternary operating bits.

\subsubsection{Sharpness-aware Learned Step Size Quantization (SALSQ)}
\label{sec:salsq}
LSQ~\cite{esser2020learned} is a technique in quantization-aware training (QAT). LSQ learns the trainable parameter $s$ using gradient-based optimization in quantization formula (2). Compared to QIL~\cite{jung2018joint}, FAQ~\cite{mckinstry2018discovering}, LQ-Nets~\cite{zhang2018lq}, PACT~\cite{choi2018pact}, and NICE~\cite{baskin2021nice}, LSQ sets the quantizer step sizes as learnable parameters to enable more flexible and adaptive quantization during training, resulting in improved performance under conditions of less than 4 bits~\cite{esser2020learned}. Despite potential issues with quantization degradation, such as those involving leaky ReLU activation functions, we suggest that any performance loss remains within acceptable boundaries. 

To minimize loss variations when transitioning from continuous to discrete weight spaces, we merge LSQ with sharpness-aware minimization (SAM)~\cite{foret2020sharpness}, introducing a new approach termed SALSQ. Given that SAM aims for uniformly low loss in neighborhoods, SALSQ effectively optimizes quantizer step sizes under the assumption of a relatively flat loss landscape. Our ablation study demonstrates the quantitative improvements, which will be presented later in Sec.~\ref{sec:ablations}.

\subsubsection{Quantized Network Inference}
\label{sec:fully_quant}
\noindent\textbf{Batch normalization folding.} In CNNs, batch normalization (BN) layers normalize channel outputs between the convolution and activation layers. During quantized model inference, BN can be fused with the previous convolution layer as an integer-type operation through BN folding~\cite{jacob2018quantization}. This technique integrates the BN layer with the preceding layer's weights and biases, boosting inference speed and efficiency.
The typical BN operation is as follows:

\vspace{-0.5em}
\begin{equation}
\hat{x}_{i}\leftarrow \frac{x_{i}-\mu_{\mathcal{B}}}{\sqrt{\sigma^2_{\mathcal{B}}+\varepsilon}}
\end{equation}

\begin{equation}
y_i \leftarrow \gamma \widehat{x}_i+\beta \equiv \mathrm{BN}_{\gamma, \beta}\left(x_i\right)
\end{equation}
\label{bn_fold}
\vspace{-0.9em}

\noindent where $\mu_{\mathcal{B}}$ and $\sigma^2_{\mathcal{B}}$ are the mean and variance of each batch, respectively. In Eq. (5), $\gamma$ and $\beta$ are learnable parameters for maintaining non-linearity. They are applied to the batch-normalized output before passing it through the activation function. Note that immediate BN folding is not feasible in our model due to the absence of convolution biases, which are only present in the two head layers for classification and bounding boxes. We intentionally omitted bias terms in the backbone to simplify calculations. This lack of convolution biases creates challenges in accommodating the offsets generated during BN folding. To address this, we introduce zero-initialized convolution biases and proceed with BN folding, thereby ensuring the process's stability, despite the minor computational cost incurred by this slight modification to the original structure.

\vspace{-0.9em}
\begin{align}
y_i &=\gamma\frac{x_{i}-\mu}{\sqrt{\sigma^2+\epsilon}}+\beta \quad (x_i=W \cdot z_{i-1}+b)\\
&=W^{\prime} \cdot z_{i-1}+b^{\prime}
\end{align}
\vspace{-0.9em}

Ultimately, the calculation is simplified as $y = W^{\prime}\cdot z_{i-1} + b^{\prime}$, where $W^{\prime}=W\frac{\gamma}{\sqrt{\sigma^2+\epsilon}}$, $z_{i-1}$ is the activation from the previous layer, and $b^{\prime}=\frac{\gamma}{\sqrt{\sigma^2+\epsilon}}(b-\mu)+\beta$. It enables the fusion of the same calculation result during inference through BN folding (or BN fusion). This involves fusing $\mu$ and $\sigma$ from EMA statistics on FP32 and $\gamma$ and $\beta$ parameters in BN with the weights and bias terms of the previous layer. With total  47 BN layers in our network architecture, BN folding drastically reduces digital computing overhead.

\begin{figure*}
    \centering
    \includegraphics[width=0.80\textwidth]{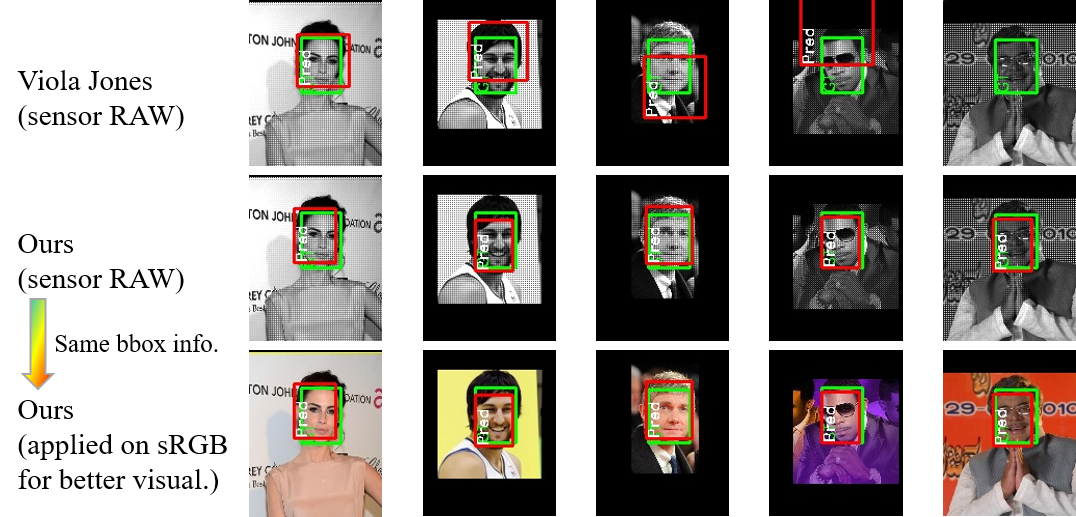}
    \caption{The qualitative comparisons on synthetic CelebCOCO. The images were rotated 90 degrees to the right to accommodate CIS device characteristics (as described in Sec.~\ref{sec:reverse_raw} `Other details.'). For better visualization, we restored the images by rotating them back to the left by 90 degrees. The third row visualizes the bounding box information from the second row (RAW image inference outputs) applied to the original (sRGB) image, solely for better visualization.}
    \label{fig_celebA_coco_output}
\end{figure*}

\begin{table*}[!htb]
\caption{The preliminary task performance and efficiency comparisons on the (sensor-RAW-like) CelebCOCO test set.}
\label{tab_CelebCOCO}
\centering
\resizebox{0.98\textwidth}{!}{%
\begin{tabular}{c|c|ccccccc|cccc}
\toprule
\multicolumn{1}{c}{Approach} & \multicolumn{1}{c}{Precision} & \multicolumn{1}{c}{AP$^{0.50}$} & \multicolumn{1}{c}{AP$^{0.75}$} & \multicolumn{1}{c}{AP$^{0.90}$} & \multicolumn{1}{c}{AP$^{S}$} & \multicolumn{1}{c}{AP$^{N}$} & \multicolumn{1}{c}{AP$^{B}$} & \multicolumn{1}{c}{\begin{tabular}[c]{@{}c@{}}False\\ Positive\end{tabular}} & \multicolumn{1}{c}{\begin{tabular}[c]{@{}c@{}}Num. of\\ Layer\end{tabular}} & \multicolumn{1}{c}{\begin{tabular}[c]{@{}c@{}}Params.\\ (M)\end{tabular}} & \multicolumn{1}{c}{\begin{tabular}[c]{@{}c@{}}FLOPs\\ (G)\end{tabular}} & \multicolumn{1}{c}{\begin{tabular}[c]{@{}c@{}}BOPs\\ (M)\end{tabular}}\\
\midrule
Viola-Jones~\cite{viola2001rapid} & FP32 & 53.8 & 2.95 & 0.15 & 57.2 & 51.9 & 50.6 & 1.36\% & - & - & - & -\\
RetinaFace~\cite{deng2020retinaface} & FP32 & 99.7 & 97.9 & 53.4 & 99.5 & 99.7 & 99.6 & 0.74\% & 138 & 0.423 & 0.050 & 2,168 \\
\midrule
Ours & FP32 & 99.7 & 96.7 & 48.3 & 99.3 & 99.6 & 99.6 & 0.81\% & 56 & 0.190 & 0.022 & 1,429\\
Ours & W4A4 & 99.3 & 93.8 & 38.4 & 98.2 & 99.1 & 98.8 & 1.06\% & 56 & 0.024 & 0.022 & 44 \\
Ours & W3A3 & 98.4 & 84.4 & 24.0 & 95.7 & 98.0 & 97.3 & 1.72\% & 56 & 0.018 & 0.022 & 32 \\
Ours (TernaryFace) & \textbf{W$_{ter}$A3} & \textbf{95.1} & \textbf{68.3} & \textbf{14.8} & \textbf{85.6} & \textbf{94.0} & \textbf{93.1} & 2.47\% & \textbf{56} & \textbf{0.012} & \textbf{0.022} & \textbf{27}\\
\bottomrule
\end{tabular}%
}
\end{table*}

\section{Experiments}
\subsection{Experimental Setup}
\noindent\textbf{Dataset.} In Sec.~\ref{sec:data_synthesis}, we used sensor-aware data synthesis techniques to create combined CelebA~\cite{liu2015faceattributes} and MS-COCO~\cite{lin2014microsoft} datasets (CelebCOCO) for training and evaluation, tailored for our main task of always-on (AO) face detection (FD) in mobile applications.
The synthetic CelebCOCO data was split into 7:2:1 proportions for training, validation, and testing, respectively. For additional research on AO FD models, we also trained and tested on the WIDER FACE benchmark dataset, which has three difficulty levels (`Easy', `Medium', and `Hard'). The WIDER FACE validation set was used for testing. Note that CelebCOCO was trained and tested at QQVGA (160$\times$120) resolution, while for the WIDER FACE dataset, we conducted training and inference at VGA (640$\times$480) resolution to ensure fair comparisons with other methods later.

\noindent\textbf{Comparison methods.} We evaluated our model performance and efficiency to analyze the quantization effect. Task performance was assessed using average precision (AP) and false positive rate, representing the ratio of bounding boxes falsely detected in scenes without faces. Efficiency evaluation considered various factors such as layers, parameter size, FLOPS, and bit operations (BOPs)~\cite{baskin2021uniq, shin2023nipq} during inference.

\noindent\textbf{Implementation details.} The model for CelebCOCO underwent training for 100 epochs with a batch size of 256. Our model, a compact version of RetinaFace with MobileNetV0.25 as the backbone, was initially trained in full precision. We progressively applied end-to-end quantization in the order of 8, 4, 3, and ternary bits, fine-tuning using initialization warm-up and a cosine annealing learning rate schedule. Model evaluation was conducted on an Intel i9-12900KF (NVIDIA RTX 3090) hardware environment, tailored for a CIS-friendly face detector optimized for sensor applications. During inference, we fused batch normalization layers, as we discussed in Sec.~\ref{bn_fold}.

\begin{figure*}[!htp]
    \centering
    \includegraphics[width=0.95\textwidth]{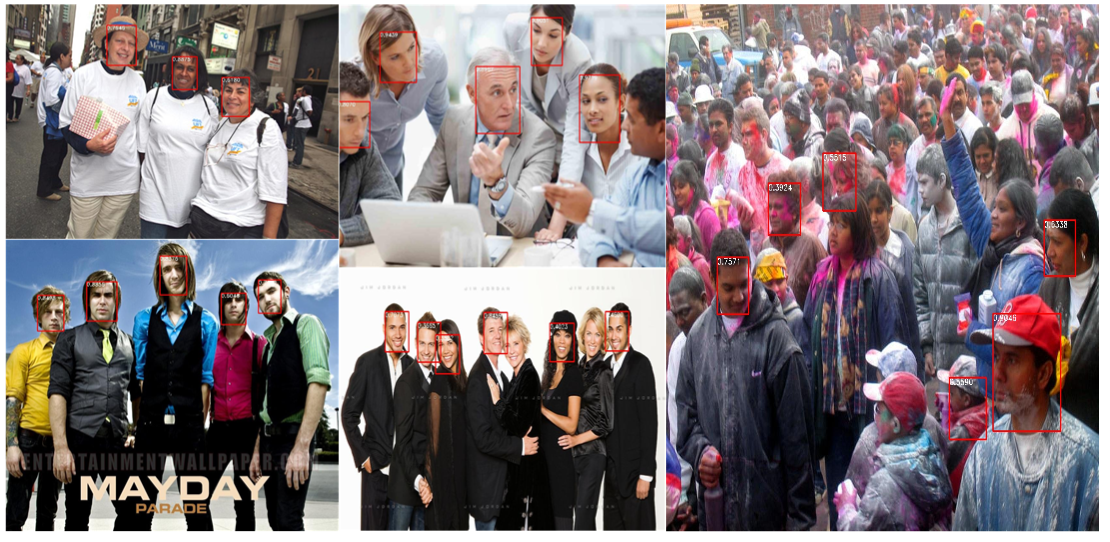}
    \vspace{-0.5em}
    \caption{The inference outputs of TernaryFace (ours) on the WIDER FACE~\cite{yang2016wider} validation set. Note that the shown images are sRGB GT images with bounding box outputs generated from a synthetic RAW-trained model's inference on synthetic RAW data. Our model is specifically designed to prioritize relatively large frontal faces, aligning with the AO FD scenario for mobile front cameras. Consequently, the model might not perform optimally on tiny faces in the Hard subset.}
    \label{fig:widerface_output}
\end{figure*}

\begin{table*}[!t]
\caption{The preliminary overall comparisons of face detection on the WIDER FACE~\cite{yang2016wider} validation set.}
\vspace{-0.5em}
\label{tab_WIDER_FACE}
\centering
\resizebox{0.75\textwidth}{!}{%
\begin{tabular}{c|c|c|ccc|ccc}
\toprule
\multicolumn{1}{c}{Approach} & \multicolumn{1}{c}{Tr/Te Inputs} & \multicolumn{1}{c}{Precision} & \multicolumn{1}{c}{Easy} & \multicolumn{1}{c}{Medium} & \multicolumn{1}{c}{Hard} & \multicolumn{1}{c}{Params.(M)} & \multicolumn{1}{c}{FLOPs(G)} & \multicolumn{1}{c}{BOPs(M)} \\
\midrule
RetinaFace & \multirow{2}{*}{\begin{tabular}[c]{@{}c@{}}RGB\\ 3ch\end{tabular}} & \multirow{2}{*}{FP32} & 88.5 & 82.8 & 55.4 & 0.423 & 0.755 & 2,168 \\
Ours &  &  & 84.1 & 75.0 & 36.0 & 0.190 & 0.339 & 1,429 \\
\midrule
RetinaFace & \multirow{6}{*}{\begin{tabular}[c]{@{}c@{}}Sensor\\ 1ch\end{tabular}} & FP32 & 75.1 & 65.8 & 36.5 & 0.423 & 0.743 & 2,168 \\
Ours &  & FP32 & 76.0 & 62.3 & 28.4 & 0.190 & 0.328 & 1,429 \\
Ours &  & W8A8 & 75.2 & 64.2 & 29.2 & 0.048 & 0.328 & 117 \\
Ours &  & W4A4 & 69.7 & 58.2 & 26.2 & 0.024 & 0.328 & 44 \\
Ours &  & W3A3 & 55.7 & 40.3 & 17.2 & 0.018 & 0.328 & 32 \\
\textbf{Ours (TernaryFace)} &  & \textbf{W$_{ter}$A3} & 32.9 & 18.6 & 7.78 & \textbf{0.012} & \textbf{0.328} & \textbf{27}\\
\bottomrule
\end{tabular}%
}
\vspace{-0.5em}
\end{table*}

\subsection{Results Considering Always-on Face Detection Scenario for CIS Mobile Applications}
In our experiment, we focused on the specific application scenario of always-on face detection in CIS for the front camera of mobile phones. We compared the performance of three face detection models: Viola-Jones (V-J)~\cite{viola2001rapid}, RetinaFace~\cite{deng2020retinaface}, and our model. The evaluation was conducted on our synthetic data, CelebCOCO test set, with a resolution of QQVGA (160$\times$120).
Additionally, we evaluated the robustness of our AO FD model by conducting inferences on three additional scenarios: (1) Small faces sub-dataset with faces sized between 4 to 10 percent of the total image, (2) Noisy faces sub-dataset with added noise for augmentation, and (3) Backlight sub-dataset with mandatory backlight augmentation. Table~\ref{tab_CelebCOCO} presents the average precision (AP) at confidence thresholds of 0.50, 0.75, and 0.90, along with the AP at confidence threshold 0.50 for the three harsh scenarios (AP$^{S}$, AP$^{N}$, AP$^{B}$) that consider real RAW capturing environments.
TernaryFace, our fully quantized network with ternary weights and 3-bit activation, demonstrated comparable performance and outperformed other models in terms of efficiency evaluation metrics. Our model boasts a compact architecture with the smallest model size (fewest parameters) and outstanding operational efficiency (lowest FLOPs and BOPs).
As discussed in Sec.~\ref{sec:data_define}, we adjusted the face size ratio in each image input to align with the AO FD scenario for front camera CIS. Our primary focus is on the average precision (AP) at a 0.50 threshold, which is enough for detecting faces in the captured images.

Output figures comparing the V-J algorithm can be seen in Figure~\ref{fig_celebA_coco_output}. Our TernaryFace model significantly outperforms V-J, a conventional method often used for hardware-level deployment as a sensor embedding algorithm. For real-world AO FD applications, TernaryFace's detection performance can be fully utilized as an always-on display unlock function for mobile phone front cameras, where V-J falls short due to its limited task performance.

\subsection{Further Works on WIDER FACE Dataset}
While our model primarily focused on the always-on face detection function in mobile phone image sensor environments, we conducted additional experiments using the benchmark dataset WIDER FACE to evaluate our model's performance. The output figures are shown in Figure~\ref{fig:widerface_output}.
To ensure fair comparison with most face detectors used with WIDER FACE, we adjusted the training and testing input image size to VGA (640$\times$480), while keeping the rest of the model architecture and data synthesis consistent with CelebCOCO. It is important to note that we fixed the number of training epochs at 300, leading to some differences in performance compared to the original RetinaFace paper~\cite{deng2020retinaface}.

As presented in Table~\ref{tab_WIDER_FACE}, our model demonstrates robust performance up to 8-bit quantization when employing lighter architectures. Interestingly, even with the inherent degradation resulting from the reverse ISP techniques used to replicate CIS RAW data, our 32-bit floating-point model slightly outperforms the original RetinaFace on the Easy subset. However, noticeable degradation becomes evident at the 4-bit condition, and a significant drop in performance occurs at the ternary-bit condition. This may be attributed to the more challenging nature of the WIDER FACE dataset, which features a greater diversity and higher quantity of faces in each scene compared to CelebCOCO.

Especially within the Hard subset, we observed a notable decline in performance, as illustrated in the rightmost image of Figure~\ref{fig:widerface_output}. In our study, we simplified the network architecture by removing the multi-scaled feature pyramid, focusing primarily on larger frontal faces. Additionally, we deliberately excluded very small faces during data synthesis and augmentation to align with scenarios involving mobile front cameras. Consequently, the model may not perform as effectively on tiny faces within the Hard subset, which is a natural result of our design decisions.

Nevertheless, our model still holds value as the first ternary face detector designed for CIS applications. It is also the first to conduct a comprehensive evaluation of quantization effects at the ternary bit-depth using WIDER FACE, the most popular dataset for face detection tasks, incorporating both sRGB and sensor-RAW levels.

\subsection{Ablation Studies}
\label{sec:ablations}
\noindent\textbf{Effect of single feature map.} The second and third rows in Table~\ref{tab_CelebCOCO} demonstrate the task performance difference between RetinaFace and our model when using only one feature map on top and eliminating the middle and bottom feature maps (as shown in Figure~\ref{fig:3_model_arch}). Despite the breakdown of the pyramidal feature map connection with residual connection to the backbone model, there is only a slight performance degradation. However, our model's performance remains competitive, especially in the AO FD task for mobile applications, with significant improvements.

\noindent\textbf{Effect of sensor RAW compared to sRGB.} In Table~\ref{tab_WIDER_FACE}, we can observe that synthetic sensor RAW data shows decreased performance compared to sRGB data in all subsets (Easy, Medium, and Hard). This study highlights the information loss caused by transitioning from 3-channel to 1-channel, as well as the significant impact of noise in the reverse ISP function on face detection performance. These results emphasize the challenges of RAW-level face detection and the need for a different approach compared to conventional face detectors.

\begin{table}[!htp]
\caption{Effect of SALSQ.}
\vspace{-0.5em}
\label{tab:compare_quant_method}
\centering
\begin{adjustbox}{width=0.85\linewidth}
\begin{tabular}{c|c|c|ccc}
\toprule
\multicolumn{1}{c}{Approach} & Precision & Method & \multicolumn{1}{c}{Easy} & \multicolumn{1}{c}{Medium} & \multicolumn{1}{c}{Hard} \\
  \midrule
\multirow{2}{*}{\begin{tabular}[c]{@{}c@{}}(Ours)\end{tabular}}
 & W4A4 & LSQ & 62.3 & 47.5 & 20.7 \\
 & W4A4 & SALSQ & \textbf{69.7} & \textbf{58.2} & \textbf{26.2} \\
\bottomrule
\end{tabular}
\end{adjustbox}
\end{table}

\noindent\textbf{Effect of SALSQ.} To assess the effect of SALSQ, we quantized our model to 4-bit and evaluated its performance on the synthetic sensor-RAW-like WIDER FACE validation set using LSQ with and without a sharpness-aware effect. The results are presented in Table~\ref{tab:compare_quant_method}. Our approach shows a significant improvement in scaling the loss gradient at the step size of each quantizer while searching the discrete spaces.

\noindent\textbf{Effect of adding the MS-COCO dataset.} When we initially conducted experiments using only the CelebA dataset, the AP at 0.5 was 98.6\%, but simultaneously, we encountered a high false positive rate of 64.84\%. 
We decided to incorporate the MS-COCO dataset, which features a more diverse range of backgrounds, with or without faces, to address the issue of forcing the detection of at least one face in each scene due to the lack of background scene diversity in CelebA. As a result, we were able to dramatically reduce the false positive rate as well as achieve the aforementioned improvements, as shown in Table~\ref{tab_CelebCOCO}.

\section{Limitation} Utilizing deep learning models for CIS may face limitations due to specialized circuits with AI accelerators, requiring comprehensive HW design validation in terms of PPA (power, performance, and silicon area) for on-sensor implementation. Additionally, varying real CIS RAW data based on manufacturers necessitates performance verification against real-world data. Despite these challenges, our fully quantized always-on face detector represents a significant step toward practical CIS applications.

\section{Conclusion}
While these are preliminary results, our proposed fully quantized ternary face detector for always-on CIS demonstrates outstanding efficiency while maintaining satisfactory performance. By employing scenario-specific data synthesis and a compact architecture with our proposed quantization methods, TernaryFace becomes a compelling option for smart image sensors in real-world mobile applications.
With a deep understanding of sensor characteristics and efficiency considerations, our energy-efficient model has the potential to significantly impact application service provisioning for CIS customers in the near future.

\section*{Acknowledgments} This work was motivated and supported in part by CIS Data Intelligence at SK hynix.
This work was supported in part by the National Research Foundation of Korea (NRF) grant funded by the Korea government (MSIT) (No. NRF-2022R1A4A1030579, NRF-2022M3C1A309202211) and Creative-Pioneering Researchers Program through Seoul National University.

{\small
\bibliographystyle{ieee_fullname}
\bibliography{egbib}
}

\end{document}